\documentclass{article}
\usepackage{hyperref}
\usepackage{amsmath,amssymb,amsfonts}
\usepackage{algorithm,tabularx}
\usepackage[noend]{algpseudocode}
\usepackage{graphicx}
\usepackage{subcaption}
\usepackage{textcomp}
\usepackage{xcolor}
\usepackage{float}
\usepackage{tikz}
\usepackage{authblk}

\usepackage{color, colortbl}
\AtBeginDocument{} % no abtract word
\title{A Semi-Dynamic Bus Routing Infrastructure based on MBTA Bus Data}
\author{Movses Musaelian, Anane Boateng, Md Zakirul Alam Bhuiyan}
\affil{Department of Computer and Information Science\\ Fordham University, NY 10458, USA}
\date{}

\begin{document}

\maketitle
\hrule
\begin{abstract}
Transportation is quickly evolving in the emerging smart city ecosystem with personalized ride sharing services quickly advancing. Yet, the public bus infrastructure has been slow to respond to these trends. With our research, we propose a semi-dynamic bus routing framework that is data-driven and responsive to relevant parameters in bus transport. We use newly published bus event data from a bus line in Boston and several algorithmic heuristics to create this framework and demonstrate the capabilities and results. We find that this approach yields a very promising routing infrastructure that is smarter and more dynamic than the existing system. 
\end{abstract}
\hfill
\hrule
% \begin{keyword}
% smart city \sep transportation analytics \sep bus system
% %\MSC[2010] 00-01\sep  99-00
% \end{keyword}

\section{Introduction}
\subsection{Background}
As traffic congestion continues growing in urban areas, more and more cities have realized that investment priority should be given to public transport modes, such as bus transit systems (BRT) instead of personal vehicles. Simply put, in congested cities, public transport modes are more efficient than personal vehicles in terms of carrying and moving people around. As city populations grow and as their economic bases shift and evolve, their housing sector adjusts, even more vehicles are expected to enter the roads each day, creating more traffic congestion. The 2012 Urban Mobility Report states that, the lack of public transportation services would have cost commuters an additional 865 million hours of delay. With growing urban population numbers, this number undoubtedly stands higher today (National Express). On average, expanding and optimizing transit services produced an economic benefit of roughly \$45 million a year by connecting urban areas in the US. There is no doubt that expanding public transportation use is key to reducing traffic congestion. One of the effective solutions to decreasing the pressure on city streets and highways is maintaining a robust and efficient public transportation system.

Public transportation in American cities has been poorly funded and provided sketchy service. According to the Washington Post\cite{Wpost}, in the past two years, nearly every US city, with a few notable exceptions, have reported a decline in their transit ridership rates. Data has shown that overall transit ridership rates hit an all-time low in 2017 since 2005, and bus ridership alone fell 5 percent. This steady decline has been attributed mostly to reliability issues and the rise of convenient alternatives such as rideshare. According to American Prospects Magazine \cite{prospectMagazine}, in New York City, there are approximately 68,000 Uber and Lyft cars, about five times as many as yellow taxis, which has caused average speeds during business hours in Manhattan’s core to drop to a crawl—about 5 to 6 miles per hour, 15 percent to 23 percent slower than in 2010, before Uber’s emergence. The convenience of alternatives has contributed to an increase in competition, leading to the upsurge of ridesharing companies such as Uber and Lyft and further increasing urban traffic congestion to a very high level. These rideshare companies are using modern technologies driven by sophisticated data analytics to gain significant market share. While most of the public transport system is still at the level of basic GPS tracking and either lack implementation of more sophisticated technologies or are very slow in adapting and implementing them. According to an annual overview of public transit usage, transit ridership fell in 31 of 35 major metropolitan areas in the US last year, including the seven cities that serve the majority of riders, with losses largely stemming from buses.

\subsection{Motivation}
In order to compete in our dynamically evolving society, there is the need for BRT to incorporate data analytics into their overall mode of operation by improving efficiency and effectiveness to optimally gain public confidence and essentially compete in the urban transportation business. In view of these problems, there is the need to find the middle-ground between the highly dynamic  ride sharing technology and the static bus routing system.

The use of big data in public bus transport system is still at its infancy. According to Welch and Widita\cite{Welch}, public transportation studies using big data began to emerge around 2013. The sources of data used in these studies are mainly GPS points and traces, smart card data, automated data such as automated passenger counts (APC), automated fare control (AFC), automated vehicle location (AVL), sensor data, mobile phone data, web data, and social media data. This type of data could produce an abundance of data daily and can be used to study anything from the behavior of individual passengers to the functioning of a large public transit system. There exist many ambiguities related to what constitutes big data, the ethical implications of big data collection and application, and how to best utilize the emerging data sets. However, within the transportation literature, there is a growing emphasis on developing sources of commonly collected public transportation data into more powerful analytical tools. A commonly held belief is that application of big data to transportation problems will yield new insights previously unattainable through traditional transportation data sets.

\subsection{Proposed Work}
Most of the previous studies have incorporated purely dynamic rideshare application models or GPS based fixed-route model. There has been limited research in the area of data-driven, semi-dynamic public transportation modeling. The lack of strategic planning and evaluation in the semi-flexible system, coupled with the lack of extensive literature review has created a huge gap in research in the public transportation industry. To bridge this gap, this research proposes to use data analytical techniques, based on existing data of a Boston bus line, to create a semi-dynamic routing framework which can improve the efficiency of that bus system whilst optimally increasing passenger satisfaction by way of increasing passenger pick-up effectiveness. 

This will be achieved by developing necessary metrics and heuristics from the data in order to achieve a responsive routing framework, which reflects the trade-off between time efficiency and boarding effectiveness, that can in turn be tuned by the bus authority. We hope to demonstrate how rudimentary data collection and analytics, which is rather non-invasive, can establish a semi-dynamic routing mechanism that is much smarter and effective than the existing static ones.

We hope also with this to contribute to the ever-growing repository of research in transportation analytics and hopefully encourage such data collection initiatives from bus transportation authorities. A data-driven method of bus schedule planning and bus allocation is demonstrated in the work. 

According to a 2009 USDOT-sponsored report, only 53\% of agencies monitor bus route performance on a monthly basis, although the survey did not distinguish between monitoring for the purpose of performance reporting and monitoring for service evaluation\cite{Mistretta}. This makes it very difficult to obtain quality data to perform insightful analysis. For decades, transportation planning analysis has consistently relied on manually collected data obtained largely through active solicitation, particularly for understanding transportation users’ behavior, e.g. household travel survey\cite{Chen}. This particular type of data, which is usually small scale and collected fairly infrequently (i.e. every 5–20 years), tends to be developed deliberately and intentionally to conduct transportation planning, evaluate specific transportation policy, and address relevant research purposes but not to improve efficiency and effectiveness. The lack of appropriate data was one of the major hurdles encountered during the data collection process for this study.

In all, we first seek to assess the latest literature and work in this field, related to such transportation analytics, to gauge the progress in this growing field. Second, we analyze the data wrangling methods and algorithms used to utilize the raw event data from the bus system. We then carefully assess some importance metrics that we calculated and used from the data such as station stopping probability, which were elemental in our routing framework. Further, we present how artificial passenger data was generate and then we dissect the routing infrastructure fueled by the respective data. Afterwards, we present a comprehensive evaluation composed of three critical components: framework responsiveness, "dry run" comparisons, and bus allocation optimization. Finally, we discuss the implications of our results and the pathway for future work and research.  
\section{Related Work}
Coleman et al.\cite{Coleman} used a data-driven approach to prioritize bus schedule revisions at New York City Transit bus network. Their approach used automatic vehicle location data and a ridership algorithm that combines automated fare collection data with other sources to infer stop-by-stop boardings and alightings for individual trips. This enabled their algorithm to pinpoint the routes most in need of schedule revisions. This process identified routes with too much capacity or running time, as well as those with too little, resource-costly schedule adjustments can be offset with resource-saving ones. Hanft et al., \cite{Hanft} utilized a fully-integrated big data sources: a neighborhood-wide analysis of performance and ridership, where 100\% data allowed planners to pinpoint specific, low-cost reroutes and stop changes to better serve riders, and identification of an optimal route split location for a long route with poor performance. This allowed for analysis throughout problem investigation as well as forecasting ridership and cost impacts of proposed service adjustments.
Chuah et al. \cite{Chuah} used taxi analytics to design and optimize bus routes.  They formulated the bus planning problem as an optimization of directed cycle graph cluster by using taxi rides dataset to determine some popular taxi rides in Singapore. From the clustered taxi rides, they filtered and select only the clusters whose commuting via existing public transport are tortuous if not unreachable door-to-door. Based on the discovered travel pattern, they proposed new bus routes that serve the passengers of these clusters.

As information technology and data collection improve, the opportunity to introduce different types of flexible transport options where demand and supply are better matched increases. There is however a need to gain a greater understanding of the technological, organizational and operational requirements that are needed within the context of a more proactive approach to mobility management which exploits the overall range of transport resource available. Wang et al.\cite{Wang} proposed a data-driven model to optimize the bus scheduling system and compared it with the existing bus scheduling system. Their model reduced the waiting time by a wide margin, which indicate the importance of data use in transportation.

Many recent papers have put forward proposals to make bus transit transport more flexible in order to compete with the more data-driven, high tech rideshare alternatives (Koffman et al.\cite{Koffman}). This has led to calls for a more flexible, efficient form of public transport. Flexible Transport Services (FTS) is an emerging term which covers services provided for passengers (and freight) that are flexible in terms of route, vehicle allocation, vehicle operator, type of payment and passenger category (Brake et al.\cite{Brake}). Flexible public transport services are neither fully demand responsive nor fixed route.  It is practiced by 39 percent of public transport operators in US (Potts et al.\cite{Potts}). Flexible transport services (FTS) have been of increasing interest in developing countries as a bridge between the use of personal car travel and fixed route transit services\cite{Ferreira}.  Quadrifoglio et al.\cite{Quadrifoglio} proposed a system called Mobility Allowance Shuttle Transit (MAST) with a concept that merges the flexibility of demand responsive transit (DRT) systems with the low cost operability of fixed-route bus systems. This MAST service has a fixed base route that covers a specific geographic zone, with a set of mandatory checkpoints with fixed scheduled departure times conveniently located at major connection points or at high density demand zones. Their results showed that the system is able to serve properly a reasonable demand while maintaining a relatively high velocity. A high level of technology is assumed to be a key element of a successful FTS\cite{Pratelli}. In conjunction with this, recent research studies are also being undertaken in advanced mathematical simulation methods to optimize the amount of ‘slack’ time required to accommodate demand-responsive service requests within the scheduled operating times\cite{Fu}. Many aspect of the planning activities of the flexible transit system still deserve significant research effort and this is particularly true for the strategic planning and evaluations phase\cite{Errico}.

Qui et al.\cite{Qiu} addressed the problem faced by transit planners by making the choice between a fixed-route policy and a flex-route policy for transit systems with a varying passenger demand, by proposing a criteria that depended on the processing of rejected requests in the assessment of the service quality function for flex-route services. They however concluded that their methodology needs reliable travel demand data and the land use plan of the service area and this  requires transit planners to make a detailed investigation before they are able to make a final decision to guarantee the transit system works as planned. Flexible Transport Services can be a promising solution for developing transport solutions, particularly in rural and remote areas where public transport is not active. Ropke and Pisinger tested heuristics on a pickup and delivery transportation problem, with time windows achieving good results in a reasonable amount of time\cite{Ropke}. The public transport Authorities can and have to influence and encourage the diffusion of IT based flexible transport systems, being able to link and optimize demand and offer of transport and also, experts working on public transport are needed for the popularization of flexible systems\cite{Ramazzotti}.

\section{Data Wrangling Methods}
A very significant component in constructing this infrastructure consisted of proper data cleaning approaches are what we could call "wrangling" methods to bring the data to an appropriate stage for application. By the term wrangling, we mean that besides the rudimentary cleaning and organization of data, it is necessary to tie events together through certain heuristics and with this, produce new metrics such as \textbf{bus lateness}, which were not present in the original event data. In this section, we aim to shed light on the steps taken and thus give a comprehensive overview of how our data pipeline worked from original raw data to results.
\subsection{Data Source}
\begin{figure}[ht]
    \centering
    \includegraphics[width=6cm]{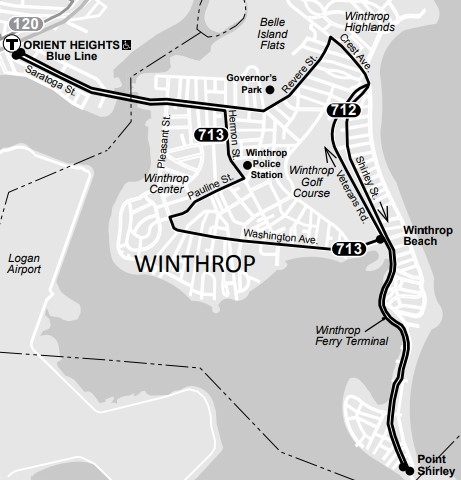}
    \caption{Map of Routes 712 and 713}
    \label{fig:routeMap}
\end{figure}

The advent of open data from city governments to public transportation has enabled researchers and enthusiasts alike to utilize such valuable data conceive and analyze better applications and systems. Our research, likewise, has benefited from this openness as MBTA (Massachusetts Bay Transportation Authority) has been diligently publishing transportation event data from their API for developers\cite{mbtaDoc}. We were interested in the bus data from the MBTA and while metro data is more abundant, the MBTA had recently begun collecting and publishing bus metric data. The only available bus data regarded two bus lines (712 and 713) as shown in Figure~\ref{fig:routeMap}, whose routes take place in Winthrop. While limited in scope, the benefit of this bus line stemmed from its geographic simplicity, which was ideal an initial infrastructure. 

The event data that we pinged from the public API gave us the necessary granularity in order to proceed with our research. Namely, we needed to know at what times the bus was arriving at the respective stations and a sense of what particular trip it belongs to. The relevant features that we used and their descriptions can be seen in Table~\ref{tb:features}.

Schedule times (in order to determine lateness) was also obtained by scraping the schedules times on the respective website. Scraping was efficient, however, schedule times are erased quite quickly (we assume the schedule times are not static for a long period of time and rather may differ or change at any day). 

\begin{table}[ht]
\renewcommand{\arraystretch}{1.3}
\caption{Relevant Features from Raw Data}
\label{tb:features}
\centering
\begin{tabular}{c||c}
\hline
\bfseries feature & \bfseries description\\
\hline \hline
timestamp & time of event\\
\hline
direction\_id & incoming or outgoing\\
\hline
event\_type & arriving or departing\\
\hline
stop\_id & id of station at which event occurred\\
\hline
trip\_id & unique distinguisher of trip within day\\
\hline
\end{tabular}
\end{table}

\subsection{Route Shortcuts}
A central part to route optimizations is applying shortcuts that can cut down on trip time given the bypassed stations are to be skipped. The process of conceiving possible shortcuts demands that we be cognizant of the streets and traffic patterns around the bus route. In our case, the geography of Winthrop yields not many opportunities for shortcuts, however, several shortcuts still exist throughout the entire bus route. 

In determining which shortcuts can be taken we had to ensure that 1) the route does indeed decrease the trip time compared to the original route 2) there are no street restrictions (e.g. one-way). Second, we had to determine how long those shortcuts would take. We believed the most accurate way was to extrapolate known trip times from similar length routes in our system and apply them for the shortcuts.

\subsection{Event Linking}
In the data wrangling process, linking events together is a careful and critical process to stitch data together rendering it for use. We had to apply event linking (described in Algorithm~\ref{alg:linking}) to our processes in order to obtain all the data components that we used for our algorithms. This linking is necessitated especially when there was missing data or errors; in essence, a sanity check for the data that we were wrangling together. It was especially useful for discarding outlier events such as if a bus broke down, which we would not want to include in our historical data aggregation.

\begin{algorithm}[ht]
\caption{Event Linking Algorithm}
\label{alg:linking}
\begin{algorithmic}[1]
\For{\text{departure in departures}}
        \State \text{min difference initialized}
        \State \text{min difference hour initialized}
\For{\text{arrival in arrivals}}
\State \text{difference = departure - arrival}
\If{\text{departure comes before arrivals}}
\If{\text{{difference $<$ minimum difference}}}
\If{\text{difference $!=$ 0}}\Comment{\textit{cannot arrive and depart at same time}}
\State \text{min difference = difference}
\State \text{min difference hour = hour(difference)}
\EndIf
\EndIf
\EndIf
\EndFor
\If{\text{minimum difference $<$ threshold}}\Comment{\textit{sanity check to exclude possible errors}}
\State \text{min difference appended to respective data list}
\EndIf
\EndFor
\end{algorithmic}
\end{algorithm}

\section{Bus Analytic Metrics}
\subsection{Bus Lateness}
\begin{figure}[ht]
    \caption{}
    \centering
    \includegraphics[width=0.90\textwidth]{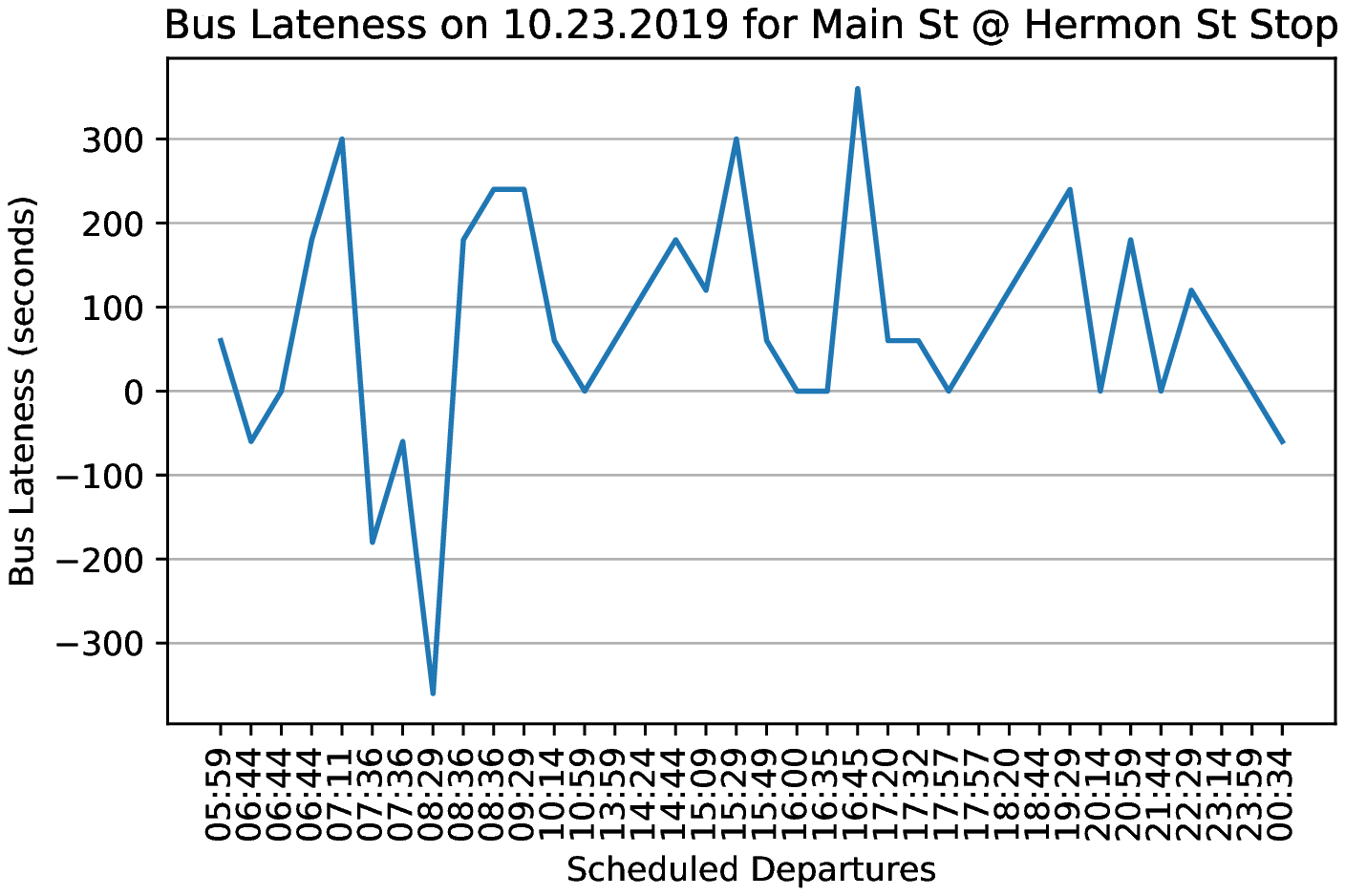}
    \label{fig:latenessexample}
\end{figure}

From a passenger point of view, bus lateness is the primary indicator of bus route performance and efficiency. This indicator was not readily available in the raw MBTA event data; thus, some work had to be done to construct this important metric. Firstly, the schedules time had to be web-scraped from the MBTA website. This was not so difficult and involved routine web-scraping techniques with the \textit{BeautifulSoup} python package. With that, the station departures had to be linked with the scheduled departures - the event linking algorithm helped link these times together. After the linkage, lateness was as expected: simply the difference between the actual departure and the scheduled departure. Figure~\ref{fig:latenessexample} shows the lateness for departures throughout a particular weekday for one of the stops on the bus route.

\subsection{Idle Time}
A routine part of bus trips is the idle time of the bus in the respective station. In other words, the difference between the departure of the bus and its arrival at a given station. In fact, idle time can be rather correlative of passenger boarding and disembarking quantities at stations, which is something not collected by the bus. Firstly though, the idle time is needed in order to incorporate into our proposed route times. We do this at a granular level of idle time per station per hour. For example, a given station may have higher idle times at rush hour when many people are boarding. The \emph{event linking algorithm} is utilized in this process to link respective events together and/or exclude data errors. In the end, we are able to use the median Idle Time for a given station for a given hour from the aggregated, linked data for our routing purposes. 

\subsection{Trip Time}
To accurately calculate our proposed route times, it is imperative to know trip times between all stations from available historical data. This computation is straight-forward as the trip time is simply the difference between the given station's arrival and the departure from the preceding station. However, missing data and data errors can hamper data linkage (e.g. the arrival at a given station belongs to a different trip coupled with the departure from its preceding station). Thus, the event linking algorithm was applied to account for such possible errors.

The second issue that had to be addressed was \textit{completely} missing data. Given our granularity of having the trip time per station pair per hour, there were some station pairs that had either no trips at certain hours or even no trips at all. The imputation method chosen was to impute the station pair's overall median trip time for hours during which there was no trip data for that station pair. For a few station pairs that had no data data at all (stations towards the end of the route), the median trip time was imputed for all their hours based on the trip time of station pairs with similar distances.

\subsection{Station Stopping Probability}
\begin{figure}[ht]
\caption{}
\centering
\includegraphics[width=0.90\textwidth]{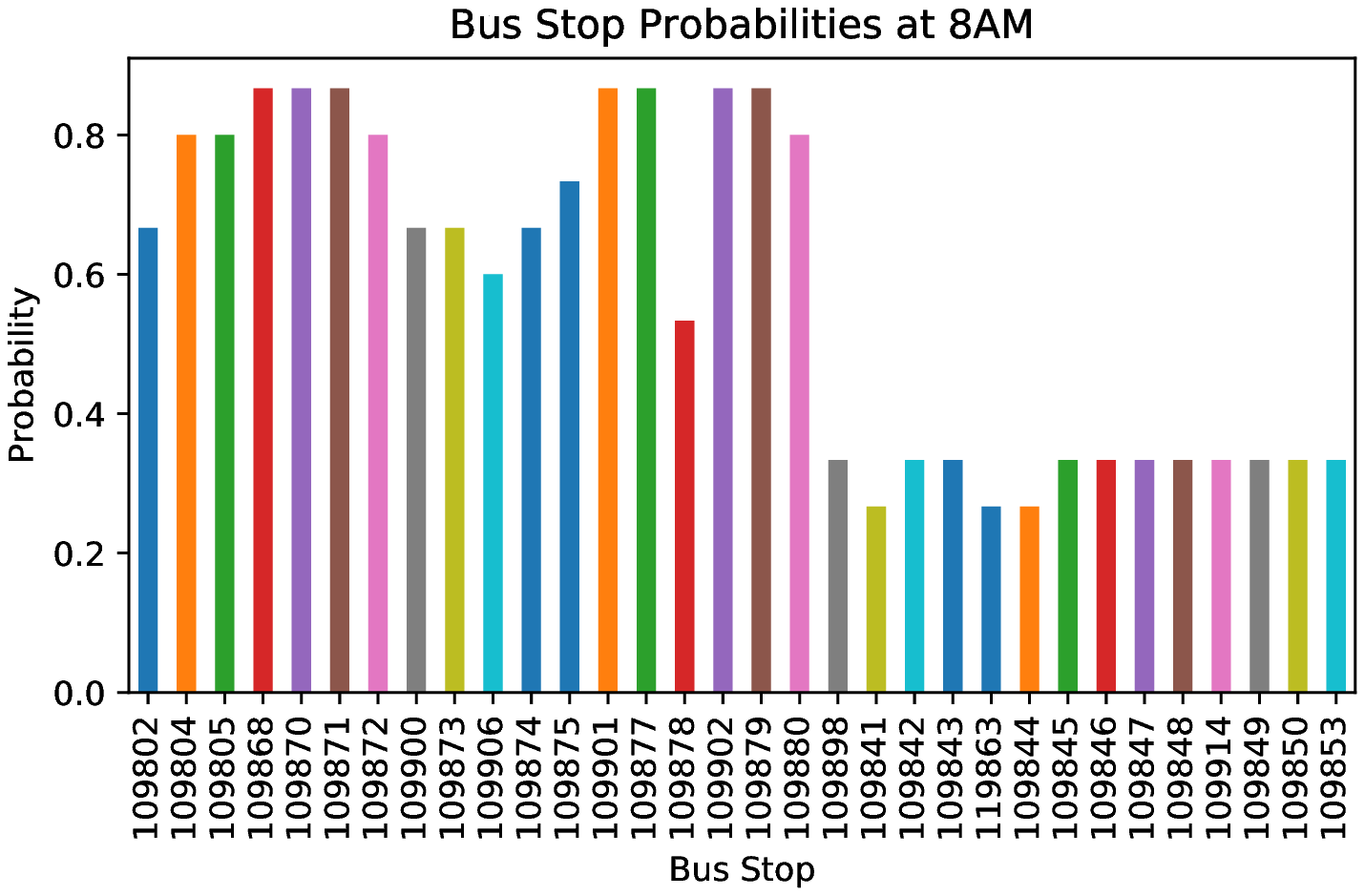}
\label{fig:stop_probability_graph}
\end{figure}

The key probability that we needed to calculate from the historical event data collected was the probability that a given bus will stop at a given station at a given hour (example shown in figure~\ref{fig:stop_probability_graph}). This would optimize our routing by providing a quantitative value for a station's "importance" given past trips. The probability is calculated with the following equation:

\begin{multline}\label{eq1}
\text{Prob\_Stop(stop\_id,hour\_of\_arrival)}= \\
 \frac{\text{\# events bus stopped at stop\_id}}
 {\text{\small{\# events bus passed through stop\_id (stopping or skipping)}}}
\end{multline}

One important heuristic is applied to determine the time at which the bus skipped a given station:
\begin{algorithm}[H]
\label{alg:skip_time}
\caption{Skipping Time Heuristic}
\begin{algorithmic}[1]
\State Find Skipped Station Time (ssTime)
\If{stop id not in trip stations}
\For{stop in preceding stops}
\If{stop id in trip stations}
\State stop\_time\_list $\longleftarrow$ stop time
\EndIf
\EndFor
\If{len(stop\_time\_list) $>0$}
\State ssTime $=$ max(stop\_time\_list)
\Else \Comment{\textit{the missed stop did not have any preceding stopped stations}}
\For{stop in succeeding stops}
\If{stop id in trip stations}
\State stop\_time\_list $\longleftarrow$ stop time
\EndIf
\EndFor
\State ssTime $=$ min(stop\_time\_list)
\EndIf
\EndIf
\end{algorithmic}
\end{algorithm}

This metric further introduces a new hyper-parameter to represent the probability threshold above which a bus on a given proposed route will stop at stations. Let us refer to this hyper-parameter as $t$. While during testing, we are able to choose different values for our $t$ as part of paramatarization, for a rudimentary baseline we can have $t$ to be set as the 25th percentile of the stopping probabilities for every given hour; thus, having a different $t$ for each hour. We can further constrain our paramatarization to be the percentile value under which stations will be skipped. We can refer to this more constrained parameter as $t_p$.  

Let us keep in mind that this probability will very likely correlate with passenger boarding at each station, which is data not available from our source. 

\subsection{Passenger Data Generation}
While our steps highlight ways that can make the route more time efficient based on shortcuts and efficient ways of station skipping; till now, we have not taken into account passenger boarding. Firstly, data on passenger boarding at each station during each trip would have been extremely useful for our work, however, such data was not available. Data that was available was the average number of boardings for each departure time (for which there is scheduled bus trip). Using this aggregate data coupled with our stopping probabilities and population density, we were able to smartly generate artificial passenger boarding data for the stations of the route at a given time. 

Our generation works as such:
\begin{algorithm}[H]
\label{alg:pass_data}
\caption{Passenger Data Generation}
\begin{algorithmic}[1]
\State probs \Comment{\textit{stopping probabilities $P_s$ for each station of the full route based on hour of proposed departure}}
\For{\text{passenger in total boardings}}
\State weighted($P_s$) random  assignment to station
\If{assigned $P_s$ is not unique} \Comment{\textit{\textbf{tiebreak} occurs with pop. density}}
\State pop. density weighted($P_s$) random  assignment to station
\Else
\State \text{original assignment}
\EndIf
\EndFor
\end{algorithmic}
\end{algorithm}

The generated passenger data gives us a very important dimension to our problem, since now we have the added optimization criteria of \emph{effectiveness}. In reality, boarding data is obviously never known before a route, however, this data gives us an important evaluation metric for how effective our route is. The trade-off between this effectiveness and time efficiency is what we need to balance in our routing framework.  

\begin{figure}[ht]
\centering
\caption{Passenger Pickup Percentages for 9:00 AM Departure after 100 simulations}
\includegraphics[width=0.90\textwidth]{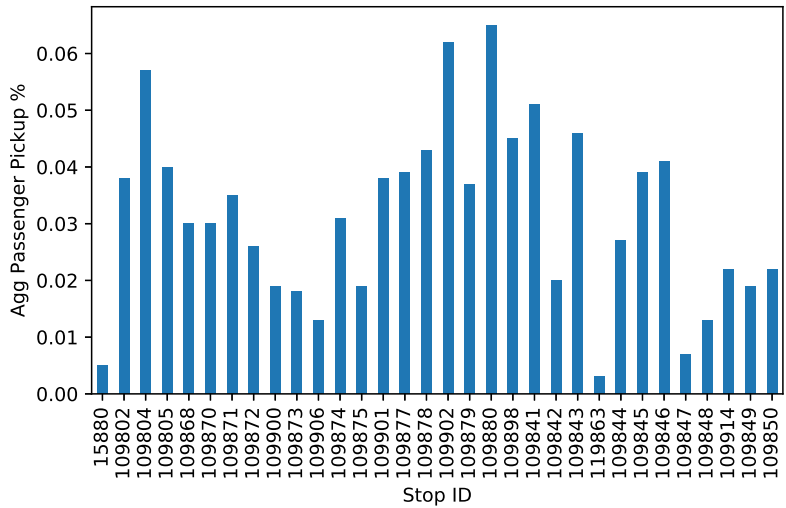}
\label{fig:pop_perct_plot}
\end{figure}

In practical application, before a route starts, the passengers waiting at stations will obviously not be known. However, historical data of passenger boardings at each station for the particular trip departure time can inform the routing. In order to replicate this historical data, we run a number of simulations (e.g. 100) of the passenger data generation and then come up with a final aggregation of the numbers to produce a "passenger pickup percentage" for each station. With this we can introduce a new parameter: Minimum Passenger Aggregation ($PA_{min}$), which denotes the minimum percentage aggregate of passenger pickup that you want to accumulate during a trip. In short, this parameter helps to optimize the route for boarding effectiveness. Figure~\ref{fig:pop_perct_plot} demonstrates this aggregate percentage for a given departure time after 100 simulations, which in practice would be 100 days of historical data for trips departing at that time. 

\section{Routing Decision Infrastructure}
Our routing infrastructure comprises of certain parameters and constraints, which can be tuned to produce optimal routes based on the needs and resources of the particular bus system. Let us itemize those parameters and constraints:
\begin{itemize}
    \item $t_p \rightarrow$ station skipping probability threshold, controls for route efficiency
    \item $PA_{min} \rightarrow$ minimum passenger aggregation, controls for route effectiveness
    \item 30 departures from origin station in a given week day
    \item Routes must be computed some period of time \emph{before} the trip day.
    \item The bus authority may want to ensure that no passenger waits more than a certain amount of time at a station. Bus allocation can be optimized for that.
    \item Particular bus has seating capacity of 36 passengers.
\end{itemize}

\begin{figure}[ht]
\caption{}
\centering
\label{fig:trip_routing_schema}
\noindent\fbox{
\parbox[50pt]{242pt}{
\textbf{If stopping at Station B:} \\
\small{$\vert Depart A \vert ---Trip Time (A \rightarrow B)--- \vert Arrive B \vert ---Idle Time B--- \vert Depart B \vert$}}}

\noindent\fbox{
\parbox[50pt]{242pt}{
\textbf{If skipping Station B:}\\
\small{$\vert Depart A \vert ---Trip Time (A \rightarrow B)--- \vert Pass B \vert ---Trip Time (B \rightarrow C)--- \vert$}}}
\end{figure}

\begin{equation}\label{eq2}
F(n) = g(n) + \sum{h(n)}\begin{cases}k \geq p, & \text{for } h_{1}(n).\\ 
k < p, & \text{for } h_{2}(n)\end{cases}
\end{equation}

\bigskip

\begin{center}\fbox{
\parbox[50pt]{240pt}{
For eq.~\eqref{eq2}, let:
\begin{itemize}
    \item $g(n)$ = total time from start to present station
    \item $h_1(n)$ = trip time + stop time from present station next station (stopping)
    \item $h_2(n)$ = trip time from present station to next station (skipping)
    \item $f(n) =$ trip time within an hour
    \item $n = 1,2,3,4,...$
\end{itemize}}}\end{center}

\bigskip

Figure \ref{fig:trip_routing_schema} shows the chronology of a trip with and without a stop. The Trip Time and Idle Time are all obtained from the historical data as described in previous sections, while equation~\eqref{eq2} shows the formulation of the routing calculation.

Given our parameters and constraints, we can implement our routing with the following steps:
\begin{enumerate}
    \item Initialize Parameters and Starting Time
    \item Calculate proposed route based on $t_p$
    \item Generate aggregate passenger boarding data for each stop (number of simulations)
    \item Revise route based on $PA_{min}$ parameter (e.g. more stations need to be added to route to full-fill minimum passenger boarding percentage). 
    \item Recalculate route arrival times if route has been changed (since previous step's revision will not result in stops being removed, shortcut check is unnecessary). 
    \item Check optimal departure of second bus based on desired minimum passenger waiting time.
\end{enumerate}

The steps of the framework as enumerated can thus serve to construct a trip based on not only specified parameters, for which we will test, but also reflecting on the historical data that is fueling all the necessary calculations for the proposed trip. 

The \textbf{second bus optimal departure} may be useful for bus authorities who want to optimize bus allocation. In general, passenger wait time can be a good indication of how efficient the bus allocation is being implemented. However, passenger wait time is very difficult to accurately measure, but we can make one assumption that wait time represents the "worst case" scenario: the passenger just misses the bus and thus has to wait for the next bus to arrive. Thus this worst case scenario waiting time is simply the arrival of second bus - departure of first bus. We can also model waiting time to be the median of the time, if the worst case is too strict. We thus employ optimal departure calculations in our results as an important logistical tool for bus authorities.

\section{Evaluation}
Our results show both the performance and perspective capability of this more dynamic framework. The historical data used for assessing all probability and trip times was limited to all the weekdays in the month of October 2019. 
\subsection{Framework Responsiveness}
\begin{figure}[ht]
\centering
\caption{Number of Stops for 7:30 AM Trip}
\includegraphics[width=.7\textwidth]{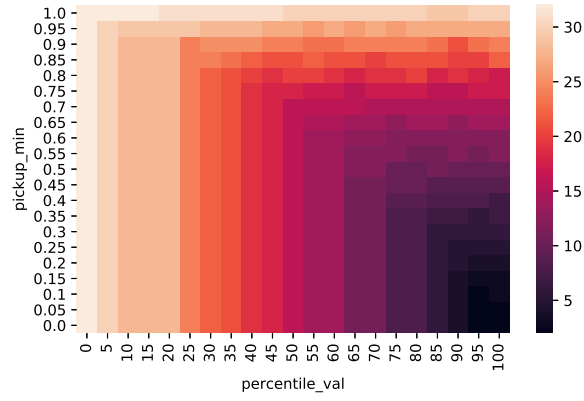}
\label{fig:num_stop_730}
\end{figure}

\begin{figure}[H]
\centering
\caption{Full Trip Times for 7:30 AM Trip}
\includegraphics[width=.7\textwidth]{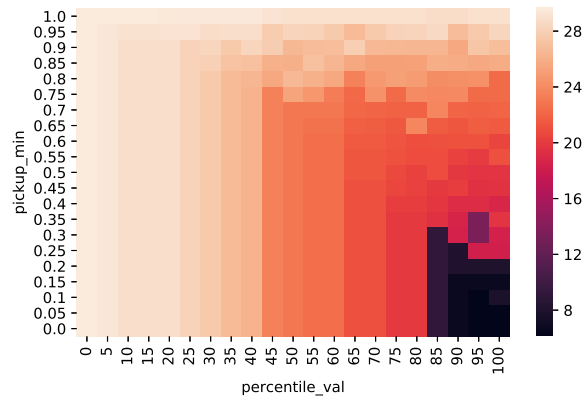}
\label{fig:trip_time_730}
\end{figure}

\begin{figure}[H]
\centering
\caption{Number of Stops for 5:00 PM Trip}
\includegraphics[width=.7\textwidth]{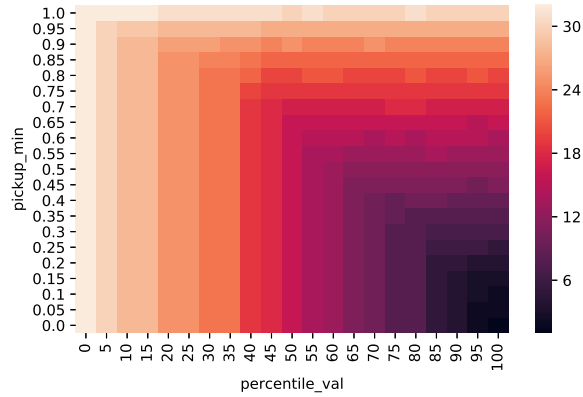}
\label{fig:num_stop_1700}
\end{figure}

\begin{figure}[H]
\centering
\caption{Full Trip Times for 5:00 PM Trip}
\includegraphics[width=.7\textwidth]{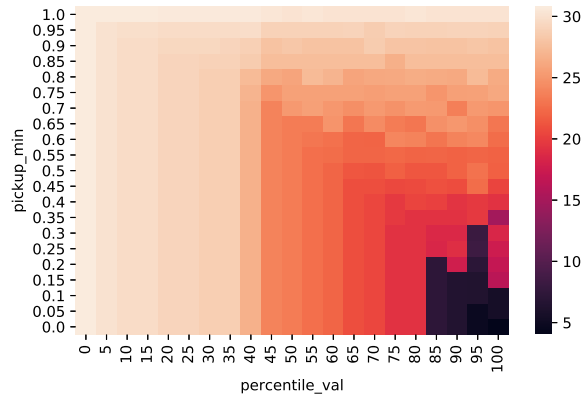}
\label{fig:trip_time_1700}
\end{figure}

It is important to assess how our routing framework reacts to differing parameter values for our two main parameters $t_p$ and $PA_{min}$ for a weekday in October. For population generation \textbf{100} simulations are run. The figures (Fig.~\ref{fig:num_stop_730}, Fig.~\ref{fig:trip_time_730}, Fig.~\ref{fig:num_stop_1700}, Fig.~\ref{fig:trip_time_1700}) for two selected departing times show the effect. 

\subsection{"Dry Run" comparison}
A way to fairly evaluate our semi-dynamic routing system to the existing static routing of the bus lines is to perform a 'dry run' for a few departure times. Given that it is not feasible to test this framework in the real-world, we must rely on simulations. Please note that we opt not to compare total trip times, simply because it would not be a fair comparison. We focus on passenger pickup as our evaluation metric. Our dry run has the following conditions:
\begin{itemize}
    \item Total passenger number inferred from available average of total passengers for given time.
    \item \textbf{100 simulations} run with passenger distr. generated in each simulation.
    \item Each route "picks up" passengers for the generated population at that time and iteration.
    \item Pick-up percentages is averaged over the 100 simulations. 
    \item Number of stops taken is recorded.
    \item Semi-Dynamic parameters may be changed. This is a natural algorithmic advantage of the semi-dynamic routing, however, the fairness of the test is not affected.
    \item Number of stops (Num\_Stops) excludes the origin station and the terminus station because those stops are mandatory (either way this doesn't affect our comparison). 
\end{itemize}

\begin{table}[ht]
\caption{Routing Dry Runs}
\label{tb:routing_dry_runs}
\centering
\begin{tabular}{|c||c||c|}
\hline 
\bfseries Type & \bfseries Pick-Up \% & \bfseries Num\_Stops\\
\hline \hline
\rowcolor{lightgray}
\multicolumn{3}{ |c| }{04.11.19 7:30}\\
\hline
Static & 0.738 & 24 \\
\hline
Semi-Dynamic & 0.924 & 25 \\
\hline
\rowcolor{lightgray}
\multicolumn{3}{ |c| }{07.11.19 13:10}\\
\hline
Static & 0.958 & 30 \\
\hline
Semi-Dynamic & 0.962 & 28 \\
\hline
\rowcolor{lightgray}
\multicolumn{3}{ |c| }{05.11.19 16:30}\\
\hline
Static & 0.954 & 30 \\
\hline
Semi-Dynamic & 0.956 & 25 \\
\hline
\rowcolor{lightgray}
\multicolumn{3}{ |c| }{07.11.19 14:40}\\
\hline
Static & 0.952 & 30 \\
\hline
Semi-Dynamic & 0.967 & 29 \\
\hline
\rowcolor{lightgray}
\multicolumn{3}{ |c| }{04.11.19 19:25}\\
\hline
Static & 0.919 & 29 \\
\hline
Semi-Dynamic & 0.967 & 29 \\
\hline
\end{tabular}
\end{table}

\subsection{Bus Allocation Optimization}
The bus authority may very well want to allocate its buses in an optimal manner. A critical metric that can reflect upon a bus system's performance is passenger waiting time at stations. Naturally, the authority will want to minimize this time to an acceptable level, which henceforth will mean that the timing of the bus allocation needs to be enhanced. A useful result from our framework has been the capability for the second bus's departure to be determined based on the set \emph{maximum median passenger waiting time}. Usually, there is a direct relationship between the set waiting time and how many minutes after the bus will depart, however, because of the semi-dynamic nature of our scheduling; this can differ based on how our routing for the particular time. In short, the process that calculates this can be summarized as such and example results are displayed in Table~\ref{tb:allocation_results}.
\begin{itemize}
    \item Record bus departure times for each station for trip A
    \item Propose trip B start 1 minute after trip A start (route semi-dynamically calculated for that start time). 
    \item See if proposed trip B arrivals times at its stations would violate the inputted median passenger waiting time. $$\frac{tripB\_arrival-tripA\_departure}{2}$$
    \item \textbf{If violated}: optimal departure for trip B is set to 1 minute before the start time that resulted in the violation.
    \item \textbf{If not violated}: Continue loop with 1 minute increments of trip B start time until there is a violation.
\end{itemize}

\begin{table}[ht]
\caption{Bus Allocation Result Examples}
\label{tb:allocation_results}
\centering
\begin{tabular}{|c||c||c|}
\hline 
\bfseries Trip A Start Time & \bfseries Trip B Start Time & \bfseries Max Wait Time\\
\hline \hline
09:30 & 09:47 & 10 \\
\hline
14:30 & 14:50 & 10 \\
\hline
15:30 & 15:59 & 15 \\
\hline
18:30 & 18:53 & 15 \\
\hline
23:15 & 23:29 & 7 \\
\hline
\end{tabular}
\end{table}

\subsection{Discussion}
Our results show the large potential that such a framework can have for making such bus routes more data driven; a welcome and much needed departure from the relatively static nature of how bus routing and scheduling is done today. First, in Fig.~\ref{fig:num_stop_730}, Fig.~\ref{fig:trip_time_730}, Fig.~\ref{fig:num_stop_1700}, and Fig.~\ref{fig:trip_time_1700}), we demonstrate how two important parameters ($t_p$ and $PA_{min}$) can be tuned based on the needs of the bus authority and what affect the parameter choices can have on the number of stops taken by the trip and the full trip times. For example, take note of how a high passenger pick-up minimum increases trip time, while a very high percentile value produces less trip times because the higher value signifies we are seeking only to stop at the most "important" stops thus reducing the stops. This trade-off interplay is what we wanted in our framework. We see a fully responsive framework that can be used to aid routing decisions based on the constraints and needs of the particular trip. With this research, we want to recognize the fact that every route can have its unique constraints and challenges, which can be most efficiently addressed with such a semi-dynamic framework.

While we focused on one particular line and direction for our research, in reality, we believe we have produced a framework general enough to apply to other routes given the rudimentary data is available. MBTA has recently begun publishing such granular bus event data and it was only for two bus lines (712 and 713). We hope as bus authorities increase the amount of data collection; it can enhance the effectiveness of such dynamic approaches and improve frameworks such as ours.

Furthermore, it must be said that it is not feasible to test our framework in the real-world given the necessary resources and permissions needed. That being said, it is important to at least have some form of evaluation to compare how our routing would do with the existing static routing. Thus, we conceived the "dry run" comparisons as the fairest way possible to conduct simulated comparisons between the two systems. While the results expectedly show our system to produce more efficient and effective routes (by pick up \% and number of stops), only real-world testing can give conclusive confidence for our system. We believe however that the dry runs provide evidence of the strength and potential of our framework and it should be highlighted that each of the runs were from 100 simulations. One metric that would have been interesting to compare would have been trip times, however, without testing this is not possible and in general, we did not use trip time as an evaluation metric given there was no way to assess how much in reality trip times would be affected. Moreover, we were using the historical trip time data from the existing data. The number of stops would be the closest metric to that. 

Finally, we present the bus allocation feature as a result of our produced framework. It shows the potential that the tool can have for bus authorities that want to more efficiently allocate buses based on a given metric; in this case, passenger waiting time. A framework that actively measures and updates the metrics that we have produced can help bus authorities make more informed decisions and have a direct view of the "health" of the given routes. Certain changes such as removing stations, increasing number of buses, moving stations, and so forth, are ones that the bus authority would make based on such data. Our framework will respond to any such change and thus inform about how effective in reality those changes were.

The granularity of our framework also affords the changes that occur throughout the year based on season and holiday. For example, maybe passenger use during the cold month of December differs from passenger use from the hotter months. Such trends can be absorbed by our algorithms; if of course, historical data is used based on month. 

In all, the results from our work have shown how the coupling of more granular bus event collection and straight-forward algorithmic analysis of that data can produce a much smarter infrastructure. Our work doesn't utilize complicated machine learning approaches, rather produces very rudimentary and essential metrics for bus routing and then creates a framework to optimize and give the authority the tools to optimize the routing so that a semi-dynamic routing can be achieved. 

\section{Conclusion \& Future Work}
In conclusion, we hope that our work has convinced readers about the potential that such a framework may have for bus transportation. Furthermore, we want to demonstrate the fact of how simple, non-invasive data collection can so significantly upgrade the data-driven capability of such a system. Our goal is not to fundamentally transform or change the essence of a bus system, but rather with rudimentary approaches make such public transportation smarter and give the bus authorities the tools to make data-driven decisions. We believe that such steps are necessary in order to give bus transportation the edge to effectively compete with ride sharing services, which offer very convenient and personalized transportation options for passengers. Furthermore, our literature review and overview of the landscape of this topic shows that serious work and thought is being given to how to make city transportation infrastructure more dynamic and data-driven in the context of data availability and the emergence of the "smart city" concept. 

We see a few potential avenues of future work. Firstly, our framework's data holds a very good use case scenario for a NoSQL database, where precomputed queries can be stored and then queried for routing algorithms. Creating such a database framework and data pipeline would help bring the framework to a more established prototype phase. Furthermore, with such a phase, it would afford us a better platform to conduct more tests with more bus lines. We especially believe that having a full year worth of the data bus data from MBTA would provide opportunity to conduct interesting seasonality analysis on our routing. For example, to see how our routing differs for warmer months versus colder months when limiting historical data to the respective seasons. In our particular bus line, we would expect that the station importance of stations near the beach during summer months would increase and thus the routing would reflect that fluctuation in probabilities. In all, this could really demonstrate how responsive the routing framework can be to seasonal patterns. 

Finally, while we were able to generate passenger data artificially which most likely correlated well with the actual passenger boarding data, it would have been a great boost for our framework if we had the actual granular passenger boarding data. Such data would really propel the strength of the framework and its capabilities. We believe that if such a capability is implemented: to record such granular passenger boarding data in a non-invasive manner (e.g. without changing existing ticketing system) it would be a great asset for both data driven routing and the "data knowledge" that could be gained from this. Most importantly, we hope our work and research can inspire and effectively contribute to the accelerating transformation of public transportation to a smarter and more informed part of the ever-changing city ecosystem. 

%\section*{References}
\bibliography{el_article}

\end{document}